\documentclass[a4paper]{article}

\usepackage{INTERSPEECH2022}
\usepackage{diagbox}
\usepackage[justification=centering]{caption}
\title{Low-resource Accent Classification in Geographically-proximate Settings: A Forensic and Sociophonetics Perspective}
\name{Qingcheng Zeng$^1$, Dading Chong$^2$, Peilin Zhou$^1$, Jie Yang$^{1,*}\thanks{$^{*}$ Corresponding Author.}$}
\address{
  $^1$School of Public Health, Zhejiang University, Hangzhou, China\\
  $^2$Peking University, China}
\email{qingchengzeng@outlook.com, 1601213984@pku.edu.cn, zhoupalin@gmail.com, jieynlp@gmail.com}

\begin{document}

\maketitle
 
\begin{abstract}
  Accented speech recognition and accent classification are relatively under-explored research areas in speech technology. Recently, deep learning-based methods and Transformer-based pretrained models have achieved superb performances in both areas. However, most accent classification tasks focused on classifying different kinds of English accents and little attention was paid to geographically-proximate accent classification, especially under a low-resource setting where forensic speech science tasks usually encounter. In this paper, we explored three main accent modelling methods combined with two different classifiers based on 105 speaker recordings retrieved from five urban varieties in Northern England. Although speech representations generated from pretrained models generally have better performances in downstream classification, traditional methods like Mel Frequency Cepstral Coefficients (MFCCs) and formant measurements are equipped with specific strengths. These results suggest that in forensic phonetics scenario where data are relatively scarce, a simple modelling method and classifier could be competitive with state-of-the-art pretrained speech models as feature extractors, which could enhance a sooner estimation for the accent information in practices. Besides, our findings also cross-validated a new methodology in quantifying sociophonetic changes.
\end{abstract}
\noindent\textbf{Index Terms}: Geographically-proximate accent classification, Accent modelling, Forensic phonetics, Sociophonetics

\section{Introduction}

Accents and dialects, with minor differences, are ubiquitous in human languages. Generally speaking, accent related research could be divided into accented speech recognition and accent classification.

Accented speech recognition usually refers to transcribing accented speech data into texts. \cite{teixeira1996accent} firstly adopted context independent Hidden Markov Model (HMM) units to classify five European accents by both accent group and gender group. In the machine learning era, various levels, ranging from acoustic, phonetic, phonotactic to prosodic features were used in machine learning models and managed to achieve lower Word Error Rate (WER). For example, \cite{biadsy2011automatic} tested different levels of features on English, Arabic, and Portuguese comprehensively. The accent classification systems before ASR systems enhanced the performance by a relatively large degree. After \cite{vaswani2017attention} and \cite{devlin2018bert}, the transformer architecture and the pretraining became the new paradigm for natural language processing (NLP) tasks \cite{chen2020adaptive,liu2021aligning} and the trend soon surged in speech technology research \cite{you2020contextualized,you2021self,xu2021semantic,chen2021self,you2022end}. Recently, \cite{li2021end} introduced an unsupervised style embedding method and witnessed 14.8\% WER reduction based on transformer architectures on AESRC2020 dataset \cite{DBLP:journals/corr/abs-2102-10233}. Generally, accented speech recognition usually models the accent as an exterior variable and merges it into downstream neural networks afterwards.
\begin{figure}[t]
  \centering
  \includegraphics[width=\linewidth]{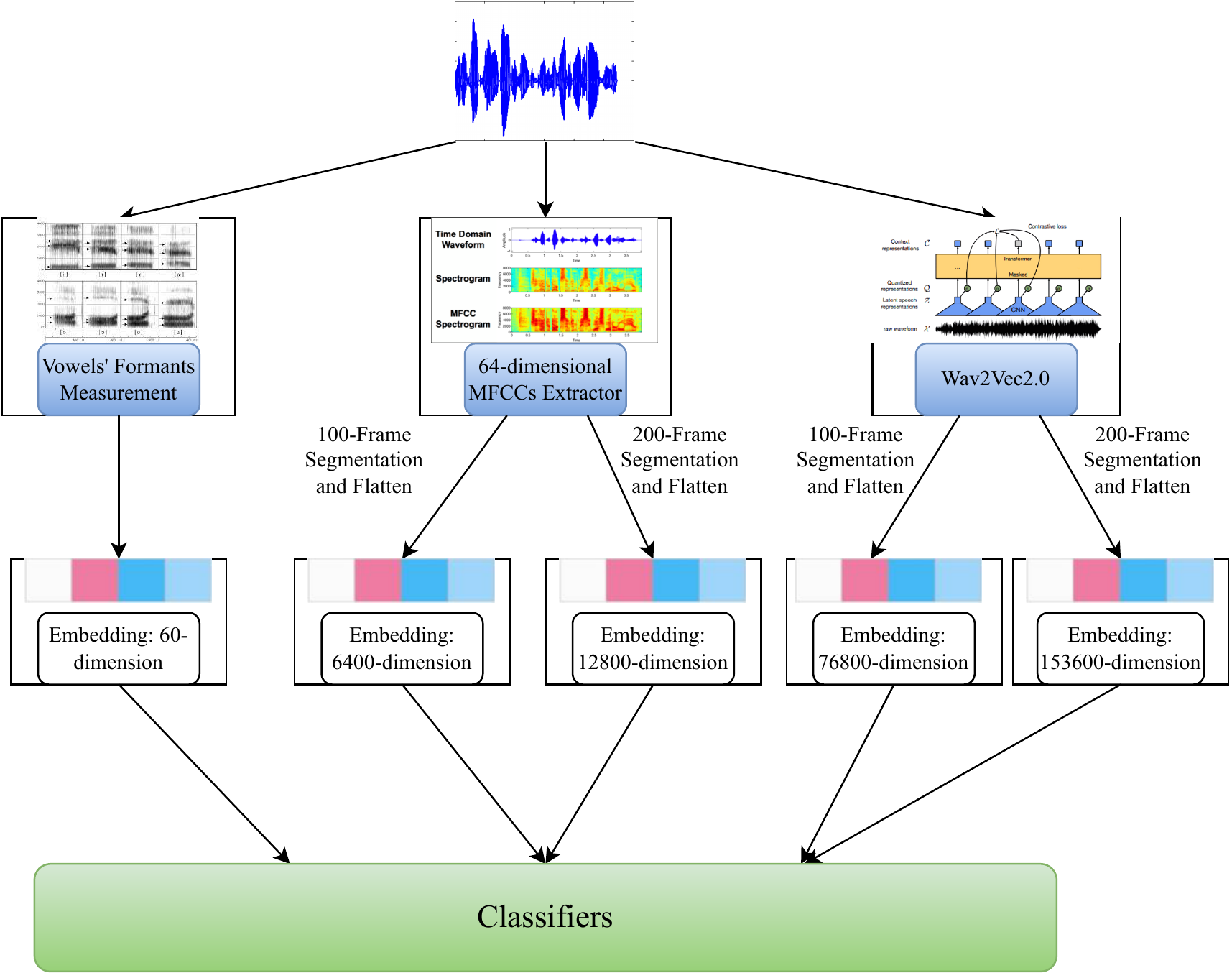}
  \caption{The Overall Binary Classification Strategy}
  \label{fig:strategy}
\end{figure}

Accent classification, which means classifying speech data into their corresponding accent categories, is relatively under-explored due to its indirect downstream applications. \cite{arslan1996language} explored foreign accented English classification with Mel-cepstrum coefficients, energy and first order differences. \cite{pedersen2007accent} tried to classify Arabic and Indian accents via Support Vector Machines (SVMs) based on MFCCs, which showed 75\% to 97.5\% accuracy with high precision and recall. \cite{guntur2022automated} compared Gaussian Mixture Models (GMM), GMM-Universal Background Model (GMM-UBM), and i-vector in classifying Dravidian accented English. \cite{tzudir21_interspeech} used convolutional neural networks (CNNs) and Gated Recurrent Units (GRUs) to classify Ao's accents with approximately 6 hours of speech.

Forensic phonetics is a rapidly growing research field whose main research tasks consist of speaker identification and disputed utterance analysis \cite{jessen2008forensic}. Accent classification is one of the most important factors in speaker identification. \cite{brown2014accdist} firstly proposed Y-ACCDIST modelling and used SVMs as the classifier to enhance accent classification in forensic applications. However, little attention was paid to forensic applications along with the development of speech technology.

Generally, speech technology in forensic applications encountered two main problems. Firstly, the scarcity of data. It is conceivable that in forensic scenarios, only short and fragmented raw speech data are available. Consequently, deep learning based methods are hard to be fully utilized. For example, \cite{rao2020improved} used 178 hours of raw speech data to improve bidirectional long short-term memory (BiLSTM) based recurrent neural networks' speech recognition performance. \cite{brown2014accdist} could only utilize about 3 hours of data to classify geographically-proximate accents. Besides, almost all accent classification research focused on foreign accented English, such as Singapore accented English or Chinese accented English. Forensic applications tend to focus more on geographically-proximate accents on the contrary.

Sociophonetics explores language variations at a social scale \cite{hay2007sociophonetics}. The methods to quantify and compute language variations are relatively new and accent classification is a window to discover machine learning models' potential in sociophonetics research.

In this paper, we mainly combined three kinds of accent modelling methods, formant measurements, MFCCs, and wav2vec2.0 with two kinds of simple classifiers, logistic regression and multiple layer perceptrons (MLPs) to explore their accent classification performances at five urban varieties in Northern England. To summarize, our main contributions are:
\begin{itemize}
    \item [1)] To the best of our knowledge, we firstly utilized pretrained speech models as feature extractors in forensic-like applications.
    \item [2)] We compared three accent modelling methods in accent classification and verified the competitiveness of traditional methods in low-resource settings.
    \item [3)] Our results suggested the feasibility of traditional classification systems as a quick screening tool in forensic practices. Besides, our results also offered a potential methodology in detecting sociophonetic variations.
\end{itemize}

\section{Data and Methods}

\subsection{Data}
The data in this paper are from \cite{strycharczuk2020general}, which were extracted from the EDAC corpus in \cite{leemann2018english}. In total, 105 speakers were selected out from the five urban varieties, Leeds (N = 27), Liverpool (N = 17), Manchester (N = 23), Newcastle upon Tyne (N = 19), and Sheffield (N = 19), among which 59\% are female and 67\% had a higher education degree. The reading essay, “The Boy Who Cried Wolf”, is a 216-word text including all English vowels. In total, 105 speakers have about 141 minutes' raw speech data. There are several advantages using this dataset. Its five adjacent urban varieties correspond well to forensic needs to classify fine-grained samples. Besides, the data were recorded via mobile phone applications which also aligns with forensic scenario characteristics.

\subsection{Experimental Setup}
The overall binary classification strategy is shown in Figure~\ref{fig:strategy}. Raw speech data are modelled by different methods and generated flattening embeddings. Accent embeddings are then processed by different classifiers to get the final results. We will discuss the experimental details in the following parts. 

\subsubsection{Feature Extraction and Segmentation}
In this paper, we used three methods based on phonetic transcription, signal processing, and pretraining respectively, which basically covered the history of speech technology research.

\textbf{Formant Measurements} Formant modelling is a traditional phonetic method which only considers vowels. Its real-world applications has been verified extensively \cite{nolan2005case}\cite{becker2008forensic}\cite{jessen21_interspeech}\cite{lo21_interspeech}. In this paper, firstly, speech data were forced-aligned via an HTK-based forced aligner \cite{Yuan2008SpeakerIO}. Then the first two formants of each vowel were measured automatically, through phonetic analysis software, Praat \cite{Boersma2022}. Different strategies were adopted for monophthongs and diphthongs. For monophthongs, the formants were extracted at midpoints. For diphthongs, onglides and offglides were defined as 20\% and 80\% of the vowel duration. Formants were extracted at onglides and offglides. After automatic measurement and hand-correction, formants' data were z-scored to reduce the effects on models’ coefficients because the second formant (F2 onwards) is bigger than the first formant (F1 onwards). In total, each speaker will be modelled by 60 features extracted from the dataset.

\textbf{MFCCs} MFCCs are one of the mostly used cepstral coefficients in signal processing, which are obtained as the inverse discrete cosine transform of the log energy in mel frequency bands. The concrete formula is shown below:

\begin{equation}
\resizebox{.9\hsize}{!}{$
  mfcc(t,c) = \sqrt{\frac{2}{M_{mfcc}}}  \sum _{m=1}^{M_{mfcc}}
  \log{\left(\widetilde{X}_m(t)\right)}
  \cos{\left(\frac{c(m-\frac{1}{2})\pi}{M_{mfcc}}\right)}
  $}
\end{equation}
where \emph{t} represents the number of frames and \emph{c} represents the dimension of MFCCs.

In this paper, MFCCs were extracted in 64 dimensions. Considering the high-dimension after modelling, we attempted to cut MFCCs into fixed number of frames, for example, 15, 100, and 200. On the one hand, data were augmented via dimensional reduction, from which we got about 45000 samples; on the other hand, this also helped to check the critical number of frames in the specific accent classification task.

\textbf{Wav2vec2.0} Wav2vec2.0 \cite{DBLP:journals/corr/abs-2006-11477} is a representative self-supervised model that outperformed semi-supervised methods in speech recognition on the Librispeech dataset \cite{7178964}. It was trained in a similar fashion to BERT, masking a fixed proportion of time steps in the latent encoder space. Besides, its performances in speaker verification and language identification are also remarkable \cite{fan21_interspeech}. In this paper, we used \emph{wav2vec-base} as our feature extractor to retrieve final layer output as speech representations for downstream classification tasks. Seemingly, we also did dimensional reduction via cutting representations into specific length of frames.

\subsubsection{Classifiers}
In this paper, we mainly used two classifiers, logistic regression and MLPs due to several reasons. Firstly, logistic regression has been tested to be one of the most robust linear classification methods. And MLPs is the simplest classifier which introduced non-linearity activation functions. Besides, both classifiers are relatively simple and can be trained in relatively quick speed, which aligns well with real-world forensic applications. At last, the results we tested on RNNs led to two intractable problems, easily overfitting and extremely slow training speed.

\subsubsection{Classification Strategy and Evaluation Metrics}
\cite{strycharczuk2020general} and \cite{tzudir21_interspeech} both trained classifiers using one-versus-all classification strategy, which we are following along in this paper. Take the training of Liverpool as an example, 10\% of the data will be selected out as the test objects. A training set with an equal number of Liverpool speakers and other cities’ speakers will be generated for each sampling. Here, other cities’ dataset is under-sampled to create a relatively balanced training set. Finally, a classifier will be trained on this new dataset, and we tested the accuracy of the test objects. This training procedure was repeated one hundred times for each sampling with differently sampled training sets to guarantee our results' robustness.

We evaluated our models' performance based on three metrics, precision (P), recall (R) and F1 score (F1).

\section{Results}

Due to the limitation of space, the binary classification results under formant modelling, MFCCs + 200-frame segmentation, and wav2vec2.0 + 100-frame segmentation are listed in Table~\ref{tab:1} to Table~\ref{tab:3} respectively. In Table~\ref{tab:4}, we listed several experimental setups' average F1 scores which have relatively good performances. 

The speech representations generated by wav2vec2.0-base, combined with 100-frame segmentation and logistic regression, performed best at 82.03 F1 score, followed by MFCCs and 200-frame segmentation at 80.88 and 81.02 F1 scores respectively. The traditional formant method have relatively mediocre performance at 73.47 and 73.54, slightly better than MFCCs and 100-frame segmentation settings. Several different numbers of the frames were tested in the experiments, 100-frame and 200-frame segmentation have best overall performance. The F1 scores' performance along with frame segmentation and the interaction between classifiers and frames will be further discussed in the next section.

The comparison across three different modelling methods showed that wav2vec2.0 performed best averagely, at 77.56 F1 score, followed by MFCCs' 75.05 and formant modelling's 73.50. Besides, wav2vec2.0 has a smaller standard deviation compared to MFCCs, which supported pretrained models' strong capacity in speech representations. However, given the fact that formant modelling did not undergo any "augmentation", its results are quite impressive as a whole.

Jumping into different cities' classification accuracy, on the one hand, there is a rather clear trend that MFCCs and wav2vec2.0 have more averagely distributed classification results than pure formant measurements. For example, Sheffield has the lowest classification accuracy in formant modelling. Among other setups, the highest accuracy for identifying Sheffield accent improved from 62.00 to 84.10, and even higher than Newcastle accent which has high accuracy previously. On the other hand, formant modelling could identify Liverpool accents incredibly well, whose accuracy is never outperformed in following experiments. These results reflect different foci in accent modelling methods.

\begin{table}[t]
\caption{Results of Formant Measurements Modelling}
\label{tab:1}
\resizebox{0.45\textwidth}{!}{%
\begin{tabular}{|c|ccc|ccc|}
\hline
Model                    & \multicolumn{3}{c|}{Logistic Regression}                        & \multicolumn{3}{c|}{MLPs}                                       \\ \hline
\diagbox{City}{Metric} & \multicolumn{1}{c|}{P}     & \multicolumn{1}{c|}{R}     & F1    & \multicolumn{1}{c|}{P}     & \multicolumn{1}{c|}{R}     & F1    \\ \hline
Leeds                    & \multicolumn{1}{c|}{68.54} & \multicolumn{1}{c|}{69.37} & 68.95 & \multicolumn{1}{c|}{69.23} & \multicolumn{1}{c|}{69.11} & 69.17 \\ \hline
Liverpool                & \multicolumn{1}{c|}{89.56} & \multicolumn{1}{c|}{91.88} & 90.70 & \multicolumn{1}{c|}{89.33} & \multicolumn{1}{c|}{89.29} & 89.31 \\ \hline
Manchester               & \multicolumn{1}{c|}{68.77} & \multicolumn{1}{c|}{72.30} & 70.49 & \multicolumn{1}{c|}{70.50} & \multicolumn{1}{c|}{70.70} & 70.60 \\ \hline
Newcastle                & \multicolumn{1}{c|}{79.20} & \multicolumn{1}{c|}{76.53} & 77.84 & \multicolumn{1}{c|}{76.21} & \multicolumn{1}{c|}{77.00} & 76.60 \\ \hline
Sheffield                & \multicolumn{1}{c|}{59.35} & \multicolumn{1}{c|}{59.42} & 59.38 & \multicolumn{1}{c|}{62.16} & \multicolumn{1}{c|}{61.84} & 62.00 \\ \hline
\end{tabular}%
}
\end{table}
\begin{table}[t]
\caption{Results of MFCCs Modelling + 200 Frames Segmentation}
\label{tab:2}
\resizebox{0.45\textwidth}{!}{%
\begin{tabular}{|c|ccc|ccc|}
\hline
Model                    & \multicolumn{3}{c|}{Logistic Regression}                        & \multicolumn{3}{c|}{MLPs}                                       \\ \hline
\diagbox{City}{Metric} & \multicolumn{1}{c|}{P}     & \multicolumn{1}{c|}{R}     & F1    & \multicolumn{1}{c|}{P}     & \multicolumn{1}{c|}{R}     & F1    \\ \hline
Leeds                    & \multicolumn{1}{c|}{69.04} & \multicolumn{1}{c|}{68.09} & 68.56 & \multicolumn{1}{c|}{80.40} & \multicolumn{1}{c|}{79.37} & 79.88 \\ \hline
Liverpool                & \multicolumn{1}{c|}{72.39} & \multicolumn{1}{c|}{75.39} & 73.86 & \multicolumn{1}{c|}{83.50} & \multicolumn{1}{c|}{85.50} & 84.49 \\ \hline
Manchester               & \multicolumn{1}{c|}{67.38} & \multicolumn{1}{c|}{63.94} & 65.61 & \multicolumn{1}{c|}{80.26} & \multicolumn{1}{c|}{83.14} & 81.67 \\ \hline
Newcastle                & \multicolumn{1}{c|}{69.10} & \multicolumn{1}{c|}{71.51} & 70.28 & \multicolumn{1}{c|}{78.82} & \multicolumn{1}{c|}{79.27} & 79.04 \\ \hline
Sheffield                & \multicolumn{1}{c|}{71.34} & \multicolumn{1}{c|}{74.33} & 72.80 & \multicolumn{1}{c|}{77.45} & \multicolumn{1}{c|}{82.72} & 80.00 \\ \hline
\end{tabular}%
}
\end{table}
\begin{table}[t]
\caption{Results of wav2vec2.0 Modelling + 100 Frames Segmentation}
\label{tab:3}
\resizebox{0.45\textwidth}{!}{%
\begin{tabular}{|c|ccc|ccc|}
\hline
Model                    & \multicolumn{3}{c|}{Logistic Regression}                        & \multicolumn{3}{c|}{MLPs}                                       \\ \hline
\diagbox{City}{Metric} & \multicolumn{1}{c|}{P}     & \multicolumn{1}{c|}{R}     & F1    & \multicolumn{1}{c|}{P}     & \multicolumn{1}{c|}{R}     & F1    \\ \hline
Leeds                    & \multicolumn{1}{c|}{80.45} & \multicolumn{1}{c|}{82.69} & 81.55 & \multicolumn{1}{c|}{71.75} & \multicolumn{1}{c|}{71.04} & 71.39 \\ \hline
Liverpool                & \multicolumn{1}{c|}{86.11} & \multicolumn{1}{c|}{86.87} & 86.49 & \multicolumn{1}{c|}{76.66} & \multicolumn{1}{c|}{85.37} & 80.78 \\ \hline
Manchester               & \multicolumn{1}{c|}{76.61} & \multicolumn{1}{c|}{78.62} & 77.60 & \multicolumn{1}{c|}{68.53} & \multicolumn{1}{c|}{72.48} & 70.45 \\ \hline
Newcastle                & \multicolumn{1}{c|}{80.20} & \multicolumn{1}{c|}{80.58} & 80.39 & \multicolumn{1}{c|}{70.52} & \multicolumn{1}{c|}{71.34} & 70.93 \\ \hline
Sheffield                & \multicolumn{1}{c|}{82.99} & \multicolumn{1}{c|}{85.25} & 84.10 & \multicolumn{1}{c|}{77.21} & \multicolumn{1}{c|}{79.58} & 78.38 \\ \hline
\end{tabular}%
}
\end{table}
\begin{table}[t]
\tiny
\caption{Different Experimental Setups' Average F1 Score under Binary Classification}
\label{tab:4}
\resizebox{0.45\textwidth}{!}{%
\begin{tabular}{|c|c|}
\hline
Experimental Setup                    & Average F1 Score \\ \hline
Formant + LR                         & 73.47            \\ \hline
Formant + MLPs                        & 73.54            \\ \hline
MFCCs + 100 Frames + LR               & 68.07            \\ \hline
MFCCs + 200 Frames + LR               & 80.88            \\ \hline
MFCCs + 100 Frames + MLPs              & 70.22            \\ \hline
MFCCs + 200 Frames + MLPs              & 81.02            \\ \hline
\textbf{wav2vec2.0 + 100 Frames + LR} & \textbf{82.03}   \\ \hline
wav2vec2.0 + 200 Frames + LR          & 76.58            \\ \hline
wav2vec2.0 + 100 Frames + MLPs         & 74.37            \\ \hline
wav2vec2.0 + 200 Frames + MLPs        & 77.26            \\ \hline
\end{tabular}%
}
\end{table}

\section{Discussion}
In this section, we will discuss accent modelling methods, machine learning classifiers, frame segmentation, and the methods' potential insights in language variation and sociophonetics.

\subsection{Accent Modelling}
The three accent modelling methods we mainly explore showed different performances in the order of wav2vec2.0, MFCCs, and formant measurements. Recalling the modelling methodology, we only measured vowels' F1s and F2s as our feature engineering output, without any consonants' information. On the contrary, wav2vec2.0 and MFCCs took the whole speech data into consideration and better results could be expected. The differences in F1 scores are mainly at cities with high accuracy and low accuracy. For example, MFCCs managed to improve Sheffield's F1 score by over 13 points. However, at the same time, the previous highest F1 score in recognizing Liverpool dropped at about 17 points. Generally, the results using wav2vec2.0 followed the similar trends to MFCCs. 

\cite{strycharczuk2020general} stated the general trend of dialect leveling at Northern England and the fact that Liverpool and Newcastle are the most distinct two accents compared to other accents. And random forests' classification results verified this finding well. In this paper, we partially verified the prior knowledge from MFCCs and wav2vec2.0's perspective. Among all experimental settings, models could classify Liverpool accents best. However, the accuracy in identifying Newcastle accents could not apparently outperform others, which indicates that Newcastle's consonantal information might hide it among other accents. Besides, formant modelling's extraordinary performance in identifying Liverpool indicated that formant modelling is quite useful in filtering accent outliers.

To sum up, our results successfully find out a general function for three accent modelling methods and suggest an elementary workflow in forensic accent classification. Firstly, we should delimit several candidate accents awaiting for classification. Then, we should consider using formant modelling to select out the most distinct accent according to requests. Afterwards, more comprehensive accents modelling methods should be adopted and classify the accents as a quick screening tool. At last, we should go into details, analyze its acoustic features accordingly to finalize our judgments.

\subsection{Classifiers}
Generally, MLPs have better performance than its counterparts, which aligns with our intuition. The only exception is the wav2vec2.0 and 100-frame segmentation setting with the highest overall performance.

As the dimensions increase, the logistic regression models' generalization ability decrease due to its linear quality. We tested logistic regression using both MFCCs and wav2vec2.0 without doing any segmentation, which leads to a over 620000-dimension vector. The average F1 scores are dropping to less than 40. Although logistic regression is quite robust in most cases, when it comes to high-dimensional computing, MLPs are a better choice with relatively stable generalization ability. The turning point between logistic regression and MLPs is basically a rule of thumb. In our settings, we can basically say 100-frame fits better to logistic regression and 200-frame fits better to MLPs. As single dimension is increasing, more advanced classifiers should be adopted.

In this paper, we only classified accents via flattening the feature vectors into linear arrays rather than high-dimensional arrays. However, sequential information and interaction is crucial in speech information. Future work should consider using CNNs and RNNs as classifiers with advanced data augmentation techniques. Besides, a comprehensive database including more geographically-proximate accents will definitely benefit training neural networks and fine-tuning pretrained speech models.

\subsection{Frame Segmentation}
\begin{figure}[t]
  \centering
  \includegraphics[width=\linewidth]{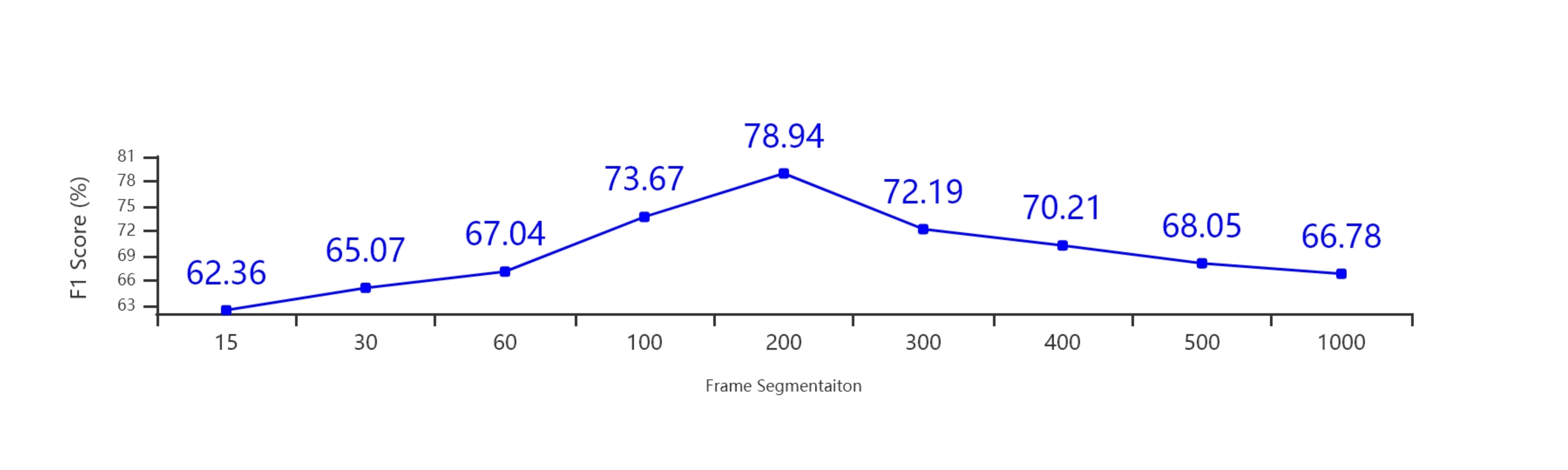}
  \caption{Average F1 Score Changes along with Different Frames of Segmentation}
  \label{fig:frame}
\end{figure}

In the results part, we only showed 100-frame and 200-frame segmentation because of their relatively high performance. Figure~\ref{fig:frame} witnessed the average F1 score changes along with different lengths' frame segmentation. Averagely, 200 frames, which approximately equals to 10 phonemes, could classify the accents best in simple classifiers' settings.

Segmenting raw speech data into fixed length of frames is a disguised way of data augmentation. Compared with formant modelling which can not make full use of data and have to encounter multicollinearity problem, segmenting according to frames both improved accuracy and generalization ability. In practices, we should consider using it as a regular way to enhance classification models.

\subsection{Sociophonetic Changes}
Quantifying sociophonetic changes automatically and discovering potential changes are rarely explored topics in computational sociolinguistics. The core difficulty lies at the trade-off between accuracy and machine learning models' interpretability. \cite{strycharczuk2020general} firstly tried to quantify sociophonetic changes via random forests. Although they argued that machine learners and human learners were following different patterns in classification, it was acceptable to use machine learning models as an exploratory tool in discovering changes. However, when it comes to logistic regression, the strong multicollinearity between formant variables made the interpretability highly fluctuant. We tried to extract the top ten most important variables from logistic regression models and compared with those from random forests. There is just one common variable. Although this variable was not seen in previous literature, further investigation is still needed to verify its real-world authenticity.

\cite{strycharczuk2020general} stated another hypothesis that the classification accuracy could reflect the degree of dialect leveling. For example, the results in Table~\ref{tab:1} implied that Liverpool and Newcastle accents are distinct ones and the rest three accents are undergoing dialect leveling. Our accent modelling methods which take overall information improved the Sheffield accent by a lot, implying that besides vowels' formants, Sheffield accents' overall information made it stand out among five accents. Previously, most variationist sociolinguistics research focused on vowels' features due to its relative scarcity and easy-to-measure acoustic features. MFCCs and pretrained models should make a comprehensive detection possible. In this paper, our experiments successfully verified the robustness of speech representations extracted from pretrained speech models, which corresponds to \cite{BARTELDS2022101137}. And our results further verified the feasibility of frame segmentation. This suggests that sociophonetic research should consider diving into fine-grained segments and pretrained models. Furthermore, we could compare across speakers in specific dimensions and even locate changes more accurately.

\section{Conclusions and Future Work}
In this paper, we explored geographically-proximate accent classification tasks, firstly utilized pretrained speech models in forensic-like settings and verified their robustness, compared across different accent modelling methods and classifiers, and proposed a general forensic accent classification workflow. We hope our work could encourage the utilization of advanced speech technology in forensic phonetics research.

As for the future work, previous computational linguistics research \cite{jawahar-etal-2019-bert} showed BERT encoded different levels of features at various layers. \cite{BARTELDS2022101137} initially showed that Transformer-based models could capture intonational and durational differences better but further investigation towards layers is still needed. Besides, pretrained models' potentials in fine-tuning have not been fully utilized. We are looking forward to seeing the enhancement of accent classification and speech representations after fine-tuning on a medium amount of speech data.

\bibliographystyle{IEEEtran}

\bibliography{mybib}


\end{document}